\documentclass[11pt,a4paper]{article}
\pdfoutput=1
\usepackage{times,latexsym}
\usepackage{url}
\usepackage[T1]{fontenc}
\usepackage{graphicx}
\usepackage{amsmath}
\usepackage{amssymb}
\usepackage{float}
\usepackage{multirow}
\usepackage{tabularx}
\usepackage[table,xcdraw]{xcolor}
\usepackage{arydshln}
\usepackage{pifont}
\usepackage{makecell}
\usepackage{adjustbox}
\usepackage{booktabs}
\usepackage{placeins}
\usepackage{caption}
\usepackage{enumitem}
\usepackage{subcaption}

\usepackage[acceptedWithA]{tacl2021v1}

\usepackage{xspace,mfirstuc,tabulary}

\newif\iftaclinstructions
\taclinstructionsfalse 
\iftaclinstructions
\renewcommand{\confidential}{}
\renewcommand{\anonsubtext}{(No author info supplied here, for consistency with
TACL-submission anonymization requirements)}
\newcommand{\instr}
\fi

\iftaclpubformat 

\else

\fi

\title{Generalizing Numerical Reasoning in Table Data through \\ Operation Sketches and Self-Supervised Learning}

\author{
  Hanjun Cho$^{1}$
  \and
  Gahyun Yoo$^{1}$
  \and
  Hanseong Kim$^{2}$
  \and
  Jay-Yoon Lee$^{1}$\thanks{Corresponding author.}
  \\
  $^{1}$Seoul National University \quad
  $^{2}$Soongsil University \\
  \texttt{\{gkswns0531, padme0421, lee.jayyoon\}@snu.ac.kr} \\
  \texttt{gkstjd1201@soongsil.ac.kr}
}

\date{}

\begin{document}
\maketitle

\newcommand{\ourmethod}{TaNOS\xspace}

\begin{abstract}
Numerical reasoning over expert-domain tables often exhibits high in-domain
accuracy but limited robustness to domain shift. Models trained with supervised fine-tuning (SFT) on specific datasets tend to rely on header–operation shortcuts rather than
structural reasoning. We introduce \textbf{TaNOS}, a continual pre-training framework comprising three components: (i) header anonymization to
reduce lexical memorization, (ii) operation sketches that provide minimal
structural cues, and (iii) self-supervised
pretraining that constructs correctness-guaranteed program–question pairs from given tables in a program-first manner. 
By decoupling domain semantics and numerical operation structure, TaNOS improves the transferability of numerical reasoning. 
Applied to an 8B instruction-tuned model, TaNOS achieves
80.13\% execution accuracy on FinQA with only 10\% train data, outperforming SFT baseline (73.97\%) with full train data and proprietary models such as GPT-5, Gemini-2.5-Pro. Furthermore, in the domain-shift experiments, TaNOS displays nearly-negligible cross-domain gap (<2pp) when standard SFT shows over 10pp gap.  
These results suggest that structural guidance with operation sketches, header-agnostic
representations, and correctness-guaranteed self-supervision can improve the robustness of numerical reasoning across diverse expert-domain tables.
\end{abstract}

\section{Introduction}

\begin{figure}[t]
\centering
\includegraphics[width=\columnwidth]{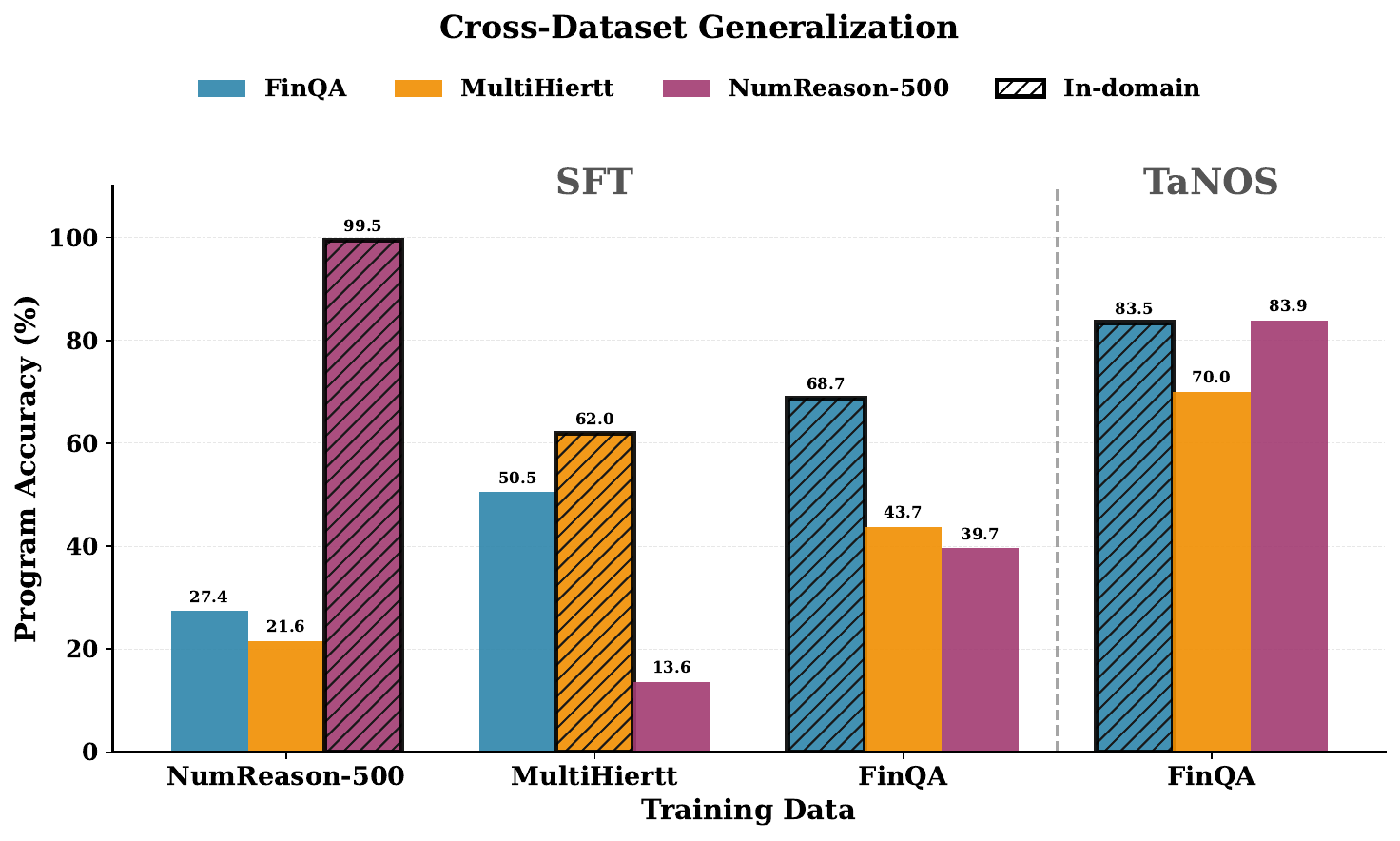}
\caption{Program Accuracy \% on \textbf{Cross-Dataset Generalization}.
Each group denotes training data; bar colors indicate test data. Hatched bars indicate in-domain evaluation; solid bars denote cross-dataset transfer. SFT degrades under subtle distribution shift.
\textbf{TaNOS}, pretrained on anonymized NumReason-500 and fine-tuned on FinQA, surpasses in-domain SFT on unseen MultiHiertt when provided with sketch guidance—indicating robust generalization.}
\label{fig:intro}
\end{figure}

Numerical reasoning over tables plays an important role in expert domains such as finance, engineering, and biology, yet remains challenging for language models~\citep{liu2023evaluating}. Unlike textual reasoning, it requires models to integrate arithmetic competence with a structural understanding of table schema, and domain-specific semantics.

A common approach is to fine-tune models on domain-specific datasets~\citep{chen2021finqa,zhu2021tat,zhao2022multihiertt,chen2022convfinqa}. While this improves in-domain accuracy, it can lead models to rely on surface-level patterns—most notably, direct associations between table headers and operations. As a result, performance often drops under shifts in schema, terminology, or question style.

Figure~\ref{fig:intro} illustrates this brittleness across three financial datasets. Despite sharing a common source (SEC reports) and exhibiting similar structures and operation types, models trained on one dataset show limited transfer to others. For example, a model trained on FinQA achieves \textbf{68.72}\% in-domain program accuracy but drops to \textbf{43.74}\% on MultiHiertt and \textbf{39.66}\% on NumReason-500 (Appendix Table~\ref{tab:cross-domain-financial}).

Through cross-domain analysis and header-shift experiments, we observe three failure modes that limit the robustness of numerical reasoning:

\begin{enumerate}
    \item \textbf{Reasoning Inefficiency.}
    Supervision signals allocate disproportionate capacity to trivial arithmetic, enabling strong performance on surface-level patterns, while generalization to unseen schemas and more complex reasoning remains limited.

    \item \textbf{Data Scarcity for Logical Supervision.}
    Numerical reasoning requires training data with consistent logical and arithmetic structure, yet standard LLM-based data generation often yields spurious or inconsistent programs, which reduces the amount of useful signal for learning more general reasoning patterns.

    \item \textbf{Header Dependency.}
    Models memorize brittle header–operation associations instead of learning relational structures. When headers change, these correlations degrade, exposing substantial lexical overfitting and limited schema generalization.
\end{enumerate}

These observations motivate the use of an \textbf{intermediate reasoning abstraction layer} that separates domain-specific lexical cues from the underlying computational structure. Our goal is to design such a layer that exposes structural supervision signals while reducing direct reliance on lexical patterns, with the aim of encouraging more domain-invariant numerical reasoning behavior.

To address this challenge, we introduce \textbf{T}able \textbf{N}umerical Reasoning with \textbf{O}peration \textbf{S}ketches (\textbf{TaNOS}), a unified framework with three complementary mechanisms designed to disentangle lexical surface forms from structural reasoning:

\begin{enumerate}
    \item \textbf{Operation Sketches.}
    We introduce minimal symbolic sketches that encode the core computational structure of a query, such as simple compositions of abstract concepts and operators (e.g. \% change of x: (x-x)/x). We assume that such sketches are available at test time, as the query itself implicitly specifies this structural information. By making these lightweight cues explicit, the model is relieved from inferring trivial arithmetic details, allowing it to devote its capacity to higher-level contextual and relational reasoning.

    \item \textbf{Self-Supervised Learning.}
    We adopt a program-first self-supervised strategy that constructs correctness-guaranteed program–question pairs from unlabeled tables. TaNOS first generates executable programs and answers, and then uses an LLM only to express them as natural-language questions, which helps maintain semantic consistency and allows training without manual annotation at scale.

    \item \textbf{Header Anonymization.}
    We propose an \textit{instance-level} header anonymization scheme that assigns unique bijective token mappings to each table. This dynamic mapping preserves structural integrity while weakening header–operation correlations.
\end{enumerate}

Together, these mechanisms are designed to address the three failure modes: operation sketches can reduce reasoning inefficiency by encouraging the model to focus more on contextual reasoning rather than obvious operation structure, self-supervised learning alleviates data scarcity for logical supervision, and instance-level header anonymization mitigates header dependency.

TaNOS unifies these mechanisms to achieve both data efficiency and robust cross-domain generalization. On a Financial TableQA benchmark, TaNOS applied to an 8B instruction-tuned model attains \textbf{85.38}\% execution accuracy with full supervision, improving over a fully supervised baseline (\textbf{73.97}\%) and 
substantially larger proprietary LLMs. With only 10\% of labeled data, TaNOS still reaches \textbf{80.13}\% accuracy, still outperforming the fully supervised and API models. In the domain-shift experiments, TaNOS maintains stable performance with negligible gap \textbf{2pp} between domains, compared to over \textbf{10pp} performance gap under SFT.

\paragraph{Contributions.}
\begin{itemize}
    \item We analyze how language models struggle with numerical reasoning 
    over expert-domain tables, identifying three failure modes: 
    reasoning inefficiency, data scarcity for logical supervision, and header dependency.
    
    \item We introduce \emph{operation sketches}, minimal structural cues that help 
    the model focus less on surface-level arithmetic and more on contextual reasoning, 
    and integrate them with header anonymization and self-supervised pretraining 
    into \textbf{TaNOS}, a unified framework achieving strong in-domain accuracy 
    and robust cross-domain generalization.
    
    \item TaNOS maintains competitive performance under limited supervision, 
    suggesting practical utility for expert domains where labeled data is scarce.
\end{itemize}

\section{Related Work}

\paragraph{Numerical Reasoning.}
Numerical reasoning involves locating supporting information from text or tables and composing operations to derive numerical answers. Early work primarily addressed text-based numerical reasoning with datasets such as DROP~\citep{dua2019drop} and MathQA~\citep{amini2019mathqa}, where models extract numerical evidence directly from passages. HybridQA~\citep{chen2020hybridqa} extends this setup by requiring reasoning over both textual and tabular contexts. To better model the structured nature of tables, encoders such as TaBERT~\citep{yin2020tabert} and TAPAS~\citep{herzig2020tapas} jointly encode cell values and headers through relational embeddings. More recently, large language models (LLMs)~\citep{chen2022large,li2023chatgpt,chen2022program} have improved general-purpose reasoning capabilities, yet numerical reasoning remains challenging~\citep{ran2019numnet,liu2023evaluating}. Most prior work focuses on general-domain settings and does not explicitly address the structural biases that arise in expert-domain tables. In contrast, our approach introduces a reasoning-level abstraction that explicitly separates surface lexical patterns from structural reasoning signals, encouraging the learning of domain-agnostic reasoning behavior.

\paragraph{Financial Numerical Reasoning.}
In the financial domain, several benchmarks evaluate reasoning over text–table pairs. FinQA~\citep{chen2021finqa,zhang2022elastic} and TATQA~\citep{zhu2021tat} require reasoning over textual context and aligned financial tables, while MultiHiertt~\citep{zhao2022multihiertt} extends this to documents with multiple hierarchical tables. Conversational variants such as ConvFinQA~\citep{chen2022convfinqa} incorporate dialogue history to model multi-turn financial analysis. Although these datasets have contributed to domain-specific numerical reasoning, they typically assume fixed schemas and stable header semantics, which may limit cross-dataset transfer. Prior methods primarily rely on supervised fine-tuning within a single dataset, which can inadvertently couple domain terminology with reasoning logic. TaNOS is intended to mitigate this limitation at the representational level by anonymizing headers and introducing operation sketches that provide abstract computational scaffolds for reasoning across schemas.

\paragraph{Self-Supervised Learning.}
Self-supervised learning (SSL) has been studied as a means of augmenting reasoning data or injecting symbolic structure into pretrained models. \citet{geva2020injecting} synthesize paired textual and numerical examples to enhance mathematical reasoning, while \citet{liu2024augment} employ SQL-based augmentation to retrieve missing information. However, most SSL approaches focus on textual or symbolic tasks and pay limited attention to structured numerical data. Our method differs in two respects: (i) it extends self-supervision to table-based reasoning by constructing correctness-guaranteed programs prior to question generation, and (ii) it uses LLMs only for controlled linguistic paraphrasing. This design combines symbolic accuracy with natural-language diversity, allowing scalable pretraining that helps preserve reasoning validity and improves generalization across domains.

\smallskip
In summary, prior work has improved architectures and datasets for numerical reasoning, but comparatively less attention has been given to abstractions that enable generalization across expert-domain tables. TaNOS contributes to this direction by integrating header anonymization, operation sketches, and correctness-guaranteed self-supervised pretraining into a unified framework.

\begin{figure*}[t]
    \centering
    \includegraphics[width=0.83\textwidth]{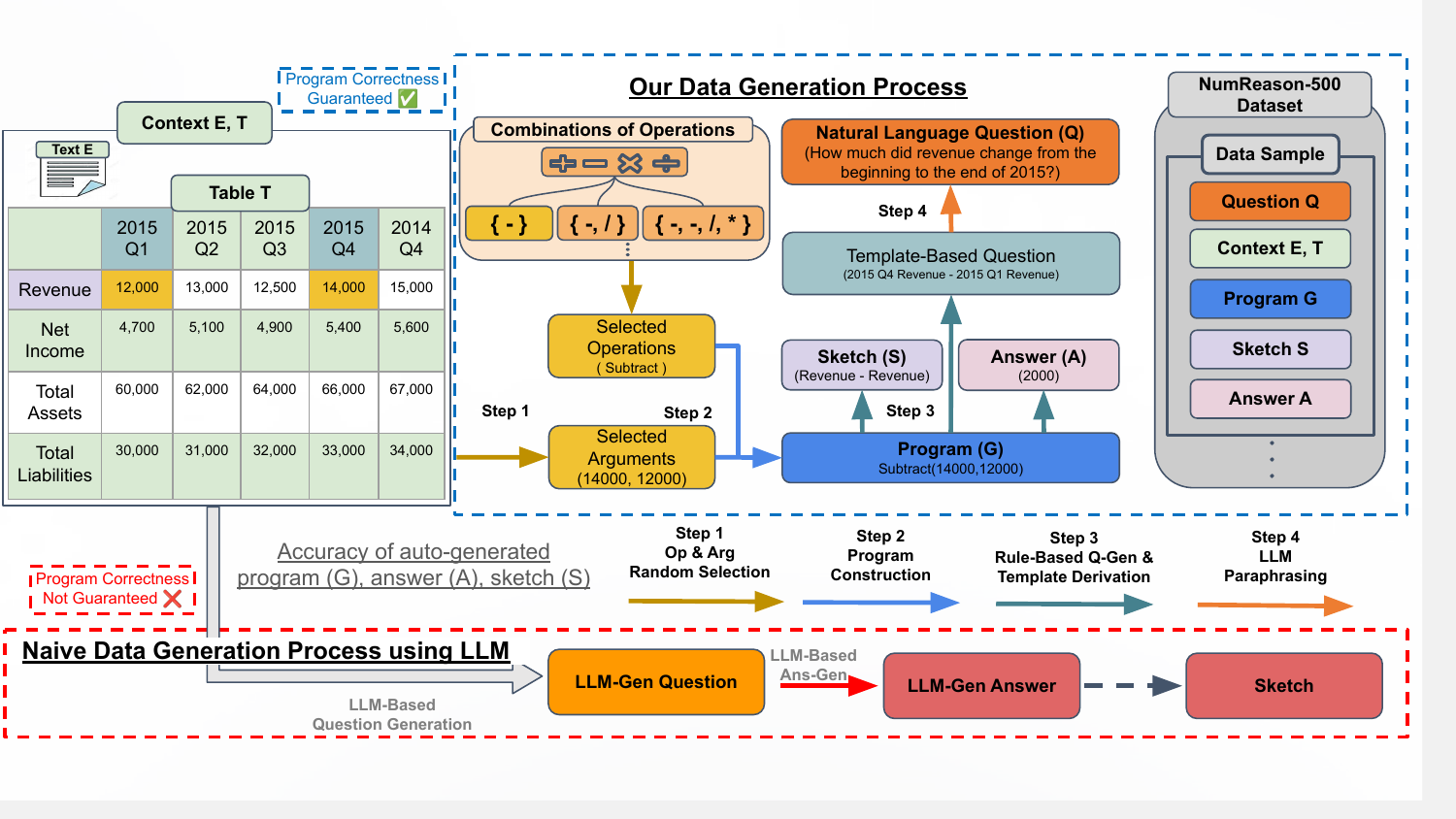}
    \caption{
    \textbf{Automatic data generation process of TaNOS.}
    Numbered steps correspond to the indicators in the figure:
    \textbf{Step 1: Operation and Argument Sampling.}
    Sample a sequence of arithmetic operations and randomly select argument cells from the table.
    \textbf{Step 2: Program Construction.}
    Combine the sampled operation sequence and arguments to form a program $G$.
    \textbf{Step 3: Program Execution and Template Derivation.}
    Execute $G$ over $(E, T)$ to obtain the answer $A$. Then extract row headers of the selected argument cells to build the operation sketch $S$, and use their row/column headers together with the operation sequence to generate a template-based question $Q$.
    \textbf{Step 4: LLM Paraphrasing.}
    Convert the template-based question into a fluent natural-language question using an LLM for paraphrasing.
    This reverse workflow (program $\rightarrow$ answer $\rightarrow$ question)
    guarantees internal consistency among $G$, $A$, and $S$ by construction (blue dashed box), 
    unlike forward-generation pipelines where correctness of the generated program, answer, and sketch is not guaranteed (red dashed box).
    }
    \label{fig:adg}
\end{figure*}

\section{Table Numerical Reasoning with Operation Sketches (TaNOS)}
\label{sec:method}

\begin{figure*}[t]
    \centering
    \includegraphics[width=\textwidth]{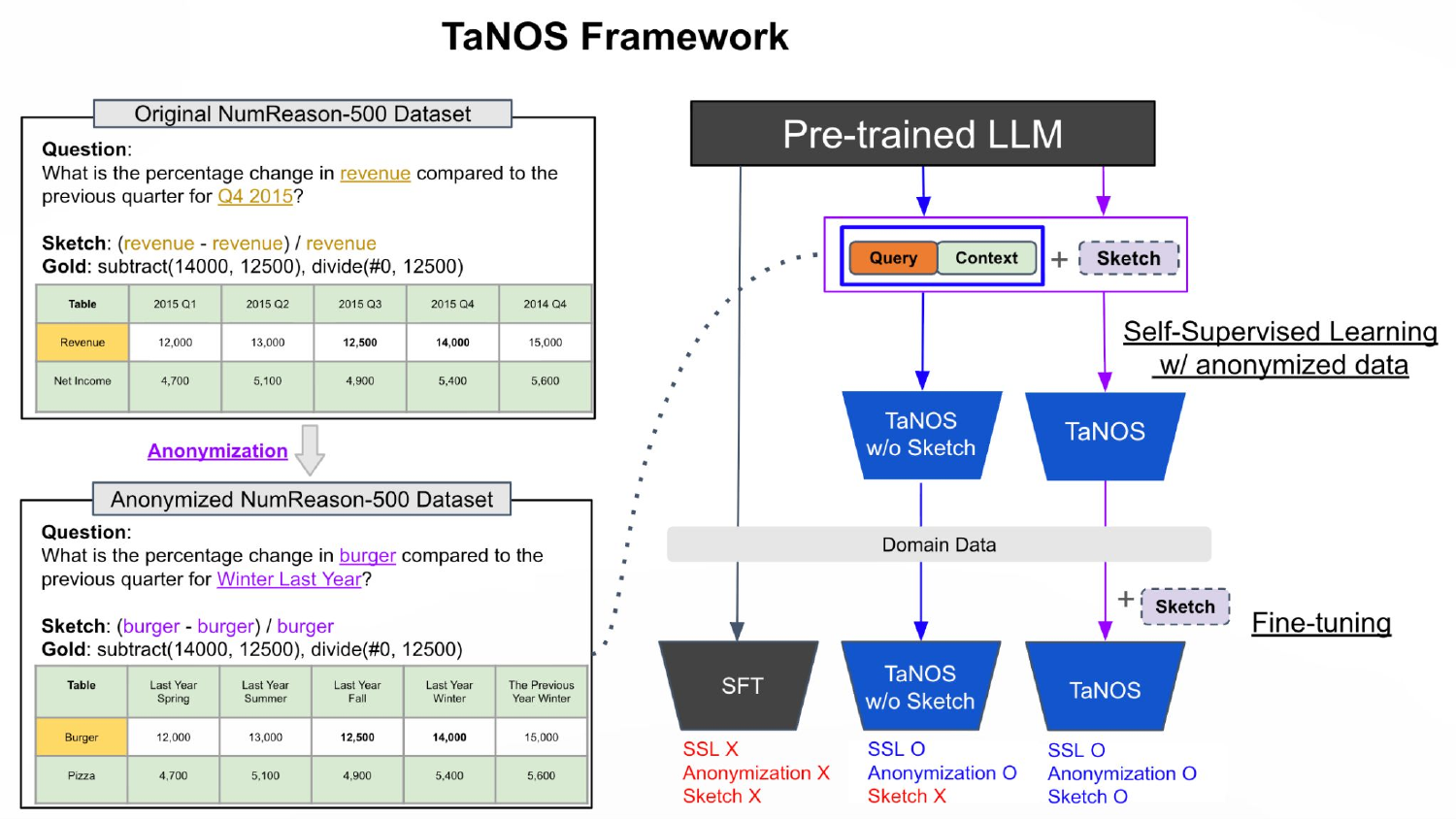}
    \caption{\textbf{Overview of TaNOS Framework.} 
    \textit{Left:} Example of instance-level header anonymization, where domain-specific headers (e.g., ``revenue'') are replaced with arbitrary tokens (e.g., ``burger''). 
    \textit{Right:} Training pipeline. Models undergo self-supervised learning on anonymized NumReason-500 with two input formats: \texttt{[Query][Context]} or \texttt{[Query][Context][Sketch]}. Fine-tuning on labeled domain data yields three system variants: SFT (no SSL), TaNOS w/o Sketch, and TaNOS.}
    \label{fig:framework}
\end{figure*}

We propose the \textbf{T}able \textbf{N}umerical Reasoning with \textbf{O}peration \textbf{S}ketches (\textbf{TaNOS}) framework,
which is designed to improve numerical reasoning in expert-domain tables through three main components:
(1) \textbf{operation sketches} that provide minimal computational guidance toward structural understanding,
(2) \textbf{header anonymization} that reduces lexical dependencies on domain-specific terms, and
(3) \textbf{self-supervised learning (SSL)} that allows large-scale pretraining without human annotation.
Each component is associated with a distinct failure mode—sketches are intended to encourage more contextual reasoning, anonymization is intended to reduce lexical bias, and SSL provides additional supervision without labels—and their combination is intended to encourage more structural, rather than surface-level, reasoning.
Figure~\ref{fig:adg} and Figure~\ref{fig:framework} illustrate the overall process.

\subsection{Operation Sketches}
\label{sec:sketches}

\paragraph{Motivation.}
Operation sketches are minimal yet interpretable computational cues that help models focus on more structural and contextual reasoning rather than memorizing specific operation patterns.
They act as lightweight scaffolds that are intended to reduce the need for the model to infer trivial arithmetic relations, and to help the model focus more on complex reasoning processes—such as contextual alignment, temporal comparison, or entity-level inference.

In TaNOS, sketches take two complementary forms:
(i) \textit{jargon-defining sketches}, which decompose domain-specific terms into explicit symbolic operations derived from rule-based templates—for example, a question asking for \emph{return on equity} corresponds to a sketch \emph{Net Income / Total Stockholders' Equity}, making the implicit formula explicit; and
(ii) \textit{computation-hint sketches}, which correspond to simple computational directions that users may naturally express when posing a question—for instance, \emph{What is the percentage change in operating cash flow from Q1 to Q2?} naturally aligns with a sketch \emph{(Operating Cash Flow - Operating Cash Flow) / Operating Cash Flow}.
These hints provide minimal guidance about the intended computation, helping the model focus on higher-level contextual reasoning rather than low-level arithmetic inference.

Sketches do not disclose the final answer or full reasoning path.
Instead, they serve as interpretable cues that are intended to help the model focus less on trivial calculation and more on abstract reasoning. (Appendix~\ref{app:numreason-examples}. Tables~\ref{tab:jargon-list} and~\ref{tab:numreason-examples})

\paragraph{Form.}
A sketch $S$ is represented as a compact symbolic sequence 
of abstract concepts and operators:
\[
S = \texttt{op}(\texttt{c}_a, \texttt{c}_b),
\]
where each $\texttt{c}$ denotes a concept token rather than 
a specific cell.
During training, sketches serve as structural blueprints 
aligning symbolic computation with tabular and linguistic 
context.

\paragraph{Inference-time availability.}
At inference time, we assume that a sketch $S$ can be provided as a lightweight specification of the intended operation (e.g., ``revenue - revenue'').
We expect this to be a relatively mild requirement, as users posing numerical questions over tables often have some sense of the intended computation, such as comparing values or computing ratios.
In our experiments, rather than relying on manual annotation, sketches are automatically extracted from the dataset using the same rule-based templates as in training (Section~\ref{sec:adgdetail}; see also Figure~\ref{fig:adg}).
To assess robustness under more realistic conditions, we additionally evaluate with paraphrased sketches that introduce lexical and syntactic variation (Section~\ref{sec:paraphrase_desc}), and find that TaNOS maintains strong performance even with noisy sketch inputs.

\paragraph{Integration into Program Generation.}
Traditional numerical reasoning models condition program generation on tables $T$, text $E$, and questions $Q$. 
TaNOS extends this by conditioning on sketches $S$:
\begin{equation*}
\begin{split}
P(A \mid T, E, Q, S)
&= \sum_{i} P(A \mid G_i, T, E, Q, S) \\
&\quad \times P(G_i \mid T, E, Q, S)
\end{split}
\end{equation*}
where $G_i$ represents an executable reasoning program and $A$ denotes the numerical answer.
The model is trained to maximize the log-likelihood of gold programs:
\[
\mathcal{L}_{\text{prog}} = 
-\sum_{t=0}^{n} \log P(w_t \mid w_{<t}, T, E, Q, S;\,\theta)
\]
By introducing sketches, TaNOS bridges basic computation and higher-level reasoning through interpretable cues,
guiding the model to capture more contextual and abstract reasoning patterns beyond surface-level arithmetic.

\begin{table*}[t]
\centering
\small
\newcommand{\val}[2]{#1\ {\scriptsize\textcolor{gray}{(#2)}}}
\resizebox{\textwidth}{!}{%
\begin{tabular}{l c | l ccccc}
\toprule
\multicolumn{2}{c|}{\textbf{LLMs (>70B)}} 
& \multicolumn{6}{c}{\textbf{LLMs (<10B)}} \\
\cmidrule(lr){1-2} \cmidrule(lr){3-8}
\textbf{Model} & \textbf{Off-the-shelf} & \textbf{Model} & \textbf{Off-the-shelf} & \textbf{SFT (10\%)} & \textbf{SFT (100\%)} & \textbf{TaNOS (10\%)} & \textbf{TaNOS (100\%)} \\
\midrule
\textit{Gemini-2.5-Pro}    & \val{76.41}{70.00}          & \textit{LLaMA-3.2-3B} & \val{10.64}{7.56}  & \val{57.56}{52.95}          & \val{69.74}{69.74}          & \val{70.38}{31.54}          & \val{82.31}{31.67} \\
\textit{Claude-4.5-Sonnet} & \val{76.79}{73.97}          & \textit{Qwen-2.5-3B}  & \val{12.44}{13.21} & \val{55.90}{59.23}          & \val{68.33}{62.69}          & \val{73.97}{43.85}          & \val{81.92}{45.64} \\
\textit{GPT-5}             & \val{\textbf{77.82}}{71.28} & \textit{Qwen-2.5-7B}  & \val{38.08}{35.90} & \val{\textbf{72.56}}{69.10} & \val{\textbf{76.15}}{75.26} & \val{79.62}{34.62}          & \val{\textbf{85.51}}{53.21} \\
\textit{Qwen-2.5-72B}     & \val{60.26}{56.54}          & \textit{LLaMA-3.1-8B} & \val{23.59}{21.15} & \val{66.67}{62.44}          & \val{73.97}{72.18}          & \val{\textbf{80.13}}{41.15} & \val{85.38}{41.03} \\
\bottomrule
\end{tabular}%
}
\caption{
\textbf{Execution Accuracy (\%) on Financial TableQA.}
The table reports execution accuracy for proprietary LLMs and instruction-tuned open-source models across five settings: off-the-shelf pretrained LLMs and fine-tuned LLMs(<10B) with SFT, TaNOS using 10\% and 100\% of the data.
By default, sketch guidance is provided at inference time, accuracy in black, and we provide extra result without sketch, indicated in gray. 
\textbf{TaNOS} is trained with sketch, which provides operation sequences,
and this training procedure encourages models to concentrate its capacity to higher-level contextual understanding of which values to retrieve from the table. As a result,
compact models (3B–8B) achieve better or competitive performance with proprietary LLMs using only 10\% of labeled data.
}
\label{tab:indomain-final}
\end{table*}

\subsection{Header Anonymization}
\label{sec:anon}

Header anonymization is intended to mitigate overfitting to memorized header tokens and to enhance generalization to unseen schemas.
Instead of a fixed mapping, TaNOS performs \textit{instance-level anonymization}, assigning distinct anonymized tokens to headers within each training sample. 
This reduces the tendency of the model to form stable header–operation associations while keeping the table structure intact.

Formally, for each instance we define a bijective mapping $\phi:\mathcal{H}\!\to\!\mathcal{V}$ assigning each header $h\!\in\!\mathcal{H}$ 
to a token $v\!\in\!\mathcal{V}$ from a reserved vocabulary range:
\[
T' = \phi(T), \quad Q' = \phi(Q), \quad S' = \phi(S).
\]
Tokens are drawn from the BERT vocabulary~\citep{devlin2018bert}, excluding special tokens.
This approach keeps the table structure intact while reducing stable lexical correlations.  
Compared to anonymization methods designed primarily for masking, 
our anonymization specifically targets header–operation correlation biases that can limit cross-domain generalization.

\subsection{Automatic Data Generation}
\label{sec:adgdetail}

To enable scalable self-supervised learning without manual labels,
TaNOS automatically constructs reasoning examples
${(T, E, Q, G, S, A)}$
from unlabeled corporate filings (SEC 10-K reports of S\&P 500 companies).
Unlike conventional pipelines that generate a question and then compute its answer,
TaNOS adopts a reverse-generation strategy—first constructing symbolic programs and answers, then generating corresponding questions (Figure~\ref{fig:adg}).
This inversion is designed to leverage the strengths of LLMs in natural-language generation while reducing their weaknesses in arithmetic reasoning, and helps ensure correctness by construction while still allowing for linguistic diversity.

\paragraph{Program and Answer Construction.}
We construct each program $G$ by sampling operation sequences of length $\ell\in[1,4]$
from an operator set $\mathcal{O}=\{\texttt{add},\texttt{sub},\texttt{mul},\texttt{div},\texttt{greater},\texttt{exp}\}$,
assigning operands based on type and unit constraints.
Executing $G$ yields an answer $A$; if execution fails or produces invalid values, we discard the example. This yields $(G, A)$ pairs whose correctness is verified by execution.

\paragraph{Question and Sketch Generation.}
Given $(T, E, G, A)$, a rule-based generator expresses each operand in $G$ using row and column headers to create canonical templates,
which are then paraphrased by an LLM to improve fluency and lexical variation while keeping $A$ fixed.
In parallel, a sketch $S$ is automatically generated by abstracting the table structure—
specifically, by discarding column headers that carry rich contextual semantics and retaining only row headers that provide simple high-level cues—
thereby capturing a high-level description of the operation pattern that is consistent with the anonymized header tokens.
All templates are rule-defined and automatically instantiated, and do not require manual effort.
After filtering invalid programs, we obtain reasoning examples, forming the \textbf{NumReason-500} dataset. (Appendix~\ref{app:numreason-examples})

\subsection{Self-Supervised Learning (SSL)}
\label{sec:ssl}

The model is pretrained on the NumReason-500 dataset using self-supervision,
treating the automatically generated questions and programs as pseudo-labels.
During pretraining, each training example is formatted as either
\texttt{[Query][Context]} or \texttt{[Query][Context][Sketch]},
depending on whether the operation sketch $S$ is included (Figure~\ref{fig:framework}).
Accordingly, we employ two pretraining configurations:
(i) \textbf{TaNOS}, which incorporates both header anonymization and operation sketches during training;
(ii) \textbf{TaNOS w/o Sketch}, which performs self-supervised pretraining on anonymized data without sketches.

To encourage active use of sketches, a subset of questions is designed to be difficult to solve without them.  
Specifically, domain-specific jargon (e.g., ``operating margin,'' ``return on equity'') is replaced with arbitrary tokens, rendering the question largely semantically opaque; the corresponding sketch thus provides the primary interpretable reasoning signal. This design simulates terminology outside the model's prior knowledge, enabling the model to handle unfamiliar domain-specific concepts through explicit structural guidance rather than memorized associations (Appendix Table~\ref{tab:jargon-list}).

LLMs are used only for natural-language paraphrasing—not for generating reasoning structures—so that the correctness of supervision is determined by the underlying programs rather than by the LLM.

Formally, let $\mathcal{D}$ denote a pretraining dataset, which depends on the configuration:
\begin{itemize}
    \item $\mathcal{D}_{\text{TaNOS}} = \{(T', E, Q', S', G)\}$: anonymized data with sketches,
    \item $\mathcal{D}_{\text{w/o Sketch}} = \{(T', E, Q', G)\}$: anonymized data without sketches,
\end{itemize}
Here, primed variables (e.g., $T'$, $Q'$) denote anonymized inputs, and $S$ denotes the operation sketch.
The training objective is:
\begin{equation*}
\mathcal{L}_{\text{SSL}} =
\mathbb{E}\left[-\log P(G \mid T, E, Q, S;\,\theta)\right],
\end{equation*}
For brevity, we write $(T, E, Q, S, G)$ generically to denote either original or anonymized inputs; when anonymization is used, $T$ and $Q$ correspond to $T'$ and $Q'$. In configurations without sketches, $S$ is omitted.

\subsection{Domain-Specific Fine-tuning}
\label{sec:finetune}

After SSL pretraining, models are fine-tuned on labeled datasets (e.g., FinQA, MultiHiertt) containing original headers and text.  
This reintroduces domain-specific language while aiming to preserve the structurally grounded reasoning learned during SSL.

\noindent
\textbf{Summary.}
TaNOS is designed to separate lexical surface information from structural reasoning.
Each component is intended to address a different aspect of the problem—anonymization to reduce lexical bias,   sketches to encourage more contextual and high-level reasoning, and SSL to provide additional supervision at scale—and their combination is intended to improve both data efficiency and cross-domain generalization in expert-domain numerical reasoning.

\section{Experiments}
\label{sec:exp}

\begin{figure*}[t!]
    \centering
    \includegraphics[width=\textwidth]{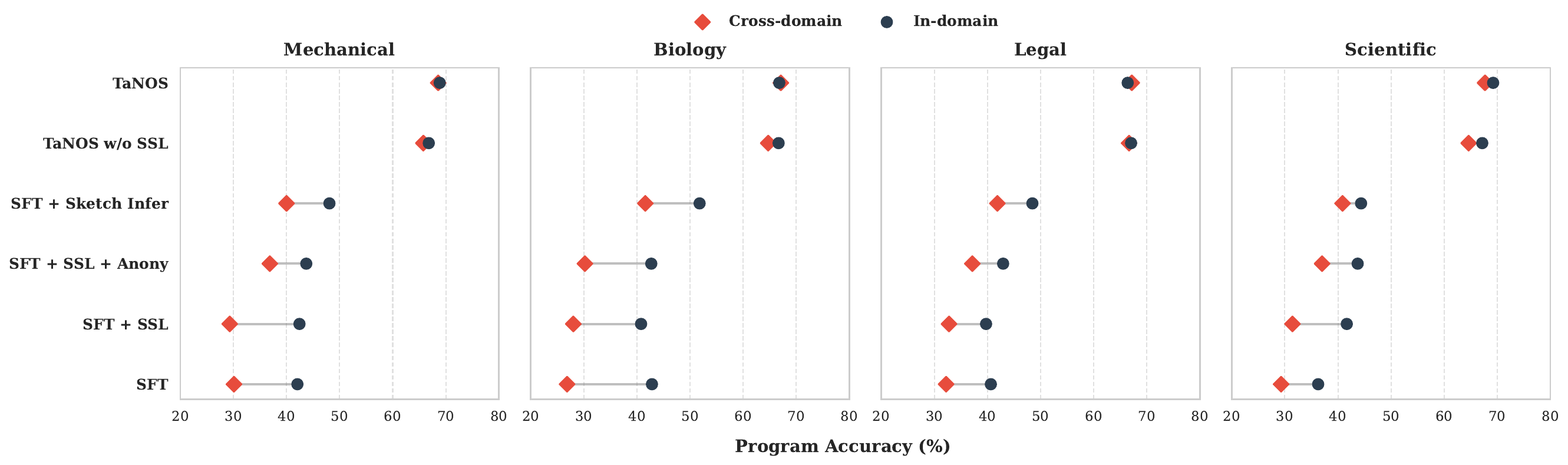}
    \caption{
        Program accuracy (\%) under \textbf{Domain Shift}. 
        Models are trained on 10\% of in-domain and evaluated in one in-domain  (blue) and four out-domains (red, averaged). Here, out-domains share the same table structure with in-domain, but have row, column headers replaced from each out-domain.
        Horizontal distances between markers indicate robustness to domain shift: \textbf{SFT} suffers large cross-domain drops, whereas \textbf{TaNOS} improves both absolute accuracy and shrinks the cross-domain gap to a nearly negligible level. Full numerical results are provided in \textbf{Appendix Table~\ref{tab:full-cross-domain-results}}.
    }
    \label{fig:cross-domain-performance}
\end{figure*}

We evaluate \ourmethod\ across multiple expert-domain numerical reasoning benchmarks to assess its impact on performance, robustness under distribution shift, and data efficiency. Our experiments are designed to answer the following questions:

\begin{enumerate}[leftmargin=*]
    \item \textbf{Overall effectiveness.} 
    Does TaNOS improve numerical reasoning performance on expert-domain tables compared to LLMs? 
    (Table~\ref{tab:indomain-final})

    \item \textbf{Component-wise robustness.} 
    How do the components of TaNOS—header anonymization, operation sketches, and self-supervised pretraining—contribute to robustness under distribution shift?
    (Figure~\ref{fig:cross-domain-performance}, Table~\ref{tab:anonymization})

    \item \textbf{Data efficiency and scalability.}
    How does TaNOS perform under varying amounts of labeled supervision, and does its data-efficiency behavior transfer across models and table benchmarks with different structural characteristics?
    (Figure~\ref{fig:data-scale}, Tables~\ref{tab:indomain-final}, ~\ref{tab:paraphrase_desc},~\ref{tab:bio_human}, and~\ref{tab:multihiertt})
\end{enumerate}

\paragraph{Financial TableQA Dataset.}
To benchmark against prior work, we use a subset of the FinQA dataset~\citep{chen2021finqa}.
This subset contains table-only questions for which operation sketches can be automatically derived by our rule-based generator. We extract 3{,}944 training, 550 development, and 780 test examples.

\paragraph{Domain-Specific Datasets.}
To evaluate cross-domain robustness, we construct four domain-shifted datasets derived from FinQA by replacing financial headers with terminology from other expert domains.
For example, the header \textit{Revenue Growth} is replaced with \textit{Tensile Strength} (Mechanical) or \textit{Cell Viability} (Biology), while maintaining the same tabular structure.
This design introduces lexical shifts in domain terminology without altering underlying reasoning patterns, which allows us to isolate the effect of lexical changes.
Each domain—\textbf{Mechanical Engineering}, \textbf{Biology}, \textbf{Legal}, and \textbf{Scientific}—contains 3{,}944 training, 550 development, and 780 test examples.

\paragraph{Automatically Generated Dataset.}
For self-supervised learning, we use the automatically generated \textbf{NumReason-500} dataset (Section~\ref{sec:adgdetail}), derived from unlabeled SEC 10-K filings of \textbf{S\&P 500} companies between 1983 and 2023.\footnote{\href{https://www.sec.gov/}{SEC filings of S\&P 500 companies.}}
During the construction of the dataset, we paraphrase all questions from rule-based templates using \textbf{GPT-3.5}, restricting the model to surface-form rewriting so that the underlying programs and answers remain unchanged.
The corpus contains 100{,}000 synthetic $(T, E, Q, G, S, A)$ tuples, where each answer $A$ is obtained by executing the program $G$. We release both \textit{original-header} and \textit{anonymized-header} versions of the dataset. For all experiments, the data are split into 70k/20k/10k train-dev-test splits. (Appendix~\ref{app:numreason-examples})
Operation sketches are included for every synthetic instance and are optionally used during fine-tuning depending on the ablation configuration (Figure~\ref{fig:cross-domain-performance}).

Together, these datasets provide a varied and controlled evaluation setting for TaNOS.
The \textbf{Mechanical Engineering}, \textbf{Biology}, \textbf{Legal}, and \textbf{Scientific} datasets are used to evaluate cross-domain robustness and general numerical reasoning behavior under varying domain semantics.
In contrast, the \textbf{NumReason-500} dataset supports systematic analysis of TaNOS's three key components---operation sketches, header anonymization, and self-supervised learning---by varying their configurations during pretraining and fine-tuning.
This dual-level design makes it possible to assess both the model's overall generalization performance and the specific contribution of each proposed mechanism.

\paragraph{Model Setup.}
We adopt the retriever–generator pipeline from FinQANet~\cite{chen2021finqa}, where a BERT-based classifier retrieves the top-3 supporting facts from linearized table rows and text sentences, which are then passed to the generator.
For the generator, we use \textbf{LLaMA-3.1-8B-Instruct}~\cite{dubey2024llama} as our \textbf{primary backbone}, and unless otherwise specified, all TaNOS variants are instantiated with this model.
This model was selected because it is widely used and has been shown to perform reliably on reasoning benchmarks, which facilitates reproducible comparisons.
To analyze the impact of model scale and to assess the applicability of TaNOS to different architectures, we also train and evaluate other models, specifically \textbf{LLaMA-3.2-3B-Instruct} and the \textbf{Qwen-2.5}~\cite{qwen2025qwen25technicalreport} series (\textbf{Qwen-2.5-3B-Instruct} and \textbf{Qwen-2.5-7B-Instruct}).

Furthermore, to contextualize TaNOS, we compare against proprietary LLMs 
(\textbf{GPT-5}~\cite{openai2025gpt5}, \textbf{Claude-4.5-Sonnet}, \textbf{Gemini-2.5-Pro}~\cite{comanici2025gemini})
and a larger open-weight model (\textbf{Qwen-2.5-72B-Instruct}, 4-bit quantized).
All open-weight models used in our experiments are \textbf{instruction-tuned} variants.
Implementation details, including hyperparameters and prompts, are provided in Appendix~\ref{app:impl}.

\paragraph{Evaluation Metrics.}
We assess model performance using two metrics: \textbf{Program Accuracy} and \textbf{Execution Accuracy}.

\textbf{Program Accuracy} measures the exact match between the predicted and gold programs, requiring identical operation sequences and arguments.
This metric directly targets the model's structural reasoning, so that correctness reflects whether the model recovers the intended reasoning path.
Since our work focuses on structured reasoning via program generation, this is the primary evaluation metric in most of our experiments.

\textbf{Execution Accuracy} evaluates the numerical correctness of the final derived answer.
For direct-answering LLMs, we parse the textual response to extract the final numerical value.
To address superficial variations—such as decimal precision, formatting differences, or insignificant rounding discrepancies—we apply standard normalization to both the predicted and gold values. (Appendix~\ref{app:impl})
A prediction is deemed correct if the normalized value is numerically equivalent to the gold answer. For program-generating models (including TaNOS), the generated program is executed to produce a deterministic result, which is then evaluated using the same protocol.

\subsection{In-domain Performance}
\label{sec:indomain}

We examine in-domain performance on the \textit{Financial TableQA}. 
Results are summarized in Table~\ref{tab:indomain-final}. 

\paragraph{TaNOS vs.\ Large Language Models.}
Across all evaluated systems, \textbf{TaNOS (100\%)} applied to the Qwen-2.5-7B backbone achieves the highest execution accuracy of \textbf{85.51\%}.
This corresponds to a gap of about 7--9 percentage points compared to proprietary models such as GPT-5, Gemini-2.5-Pro, and Claude-4.5-Sonnet, even when they are given operation sketches at inference time.
The compact LLaMA-3.2-3B model also reaches \textbf{82.31\%} with TaNOS, which is comparable to or higher than the proprietary systems in our evaluation.
These results suggest that small-scale open models can perform competitively with closed-source LLMs on specialized numerical reasoning tasks.
Collectively, these findings indicate that TaNOS can improve structural numerical reasoning over expert-domain tables. (Appendix Table~\ref{tab:appendix-llm-exec-acc})

\paragraph{Sketches as training signals rather than prompts.}
Operation sketches yield consistent but modest improvements when used as inference prompts for general LLMs.
For instance, Qwen-2.5-72B gains about 3.72pp, and proprietary models such as GPT-5 (6.54pp) and Claude-4.5-Sonnet (2.82pp) show similarly limited improvements.
The SFT baseline for LLaMA-3.1-8B reaches 73.97\% when provided with sketches.
In contrast, TaNOS (100\%) uses these sketches as structural training signals and achieves 85.38\%, about 11 percentage points higher than the SFT baseline.
This indicates that our proposed training of TaNOS with sketches allows the model to dedicate its capacity to higher-order contextual reasoning of which values to retrieve from the table, rather than allocating capacity to infer operations from scratch.
As a byproduct, TaNOS becomes dependent on sketch guidance at inference time: 
without it, performance drops to 41.03\%. 
In typical use cases, however, users formulating numerical questions 
are likely to already have an implicit understanding of the intended operations 
(e.g., sum, average, difference), making this requirement relatively lightweight.

\paragraph{Data efficiency and utility in privacy-constrained environments.}
\ourmethod\ improves data efficiency. With 10\% of the labeled data, the LLaMA-3.1-8B model reaches 80.13\%. This is higher than the \textbf{SFT (100\%)} baseline (73.97\%) and the \textbf{GPT-5} (77.82\%), even when both are provided with operation sketches. This setting is particularly relevant for privacy-sensitive sectors such as finance, where regulatory constraints can limit the use of external cloud-based APIs and encourage the deployment of local models. These results suggest that \ourmethod\ can provide competitive reasoning performance in such local deployments, using a relatively compact model even when labeled data are limited.

\subsection{Component-wise Analysis of Cross-domain Generalization}
\label{sec:cross-domain}

We evaluate cross-domain generalization on four domain-shifted variants of the Financial TableQA benchmark: Mechanical, Biology, Legal, and Scientific. The overall trends are summarized in Figure~\ref{fig:cross-domain-performance}, and detailed numerical results for all train–test domain pairs are provided in Appendix Table~\ref{tab:full-cross-domain-results}. For each training domain, we report average accuracy across all evaluation domains (\textbf{Avg.}), accuracy on the training (in-domain) domain (\textbf{In-domain}), and the mean over the remaining target domains (\textbf{Cross-domain}).

\paragraph{Gaps under standard supervised fine-tuning.}
Purely \textbf{SFT} shows a systematic gap between in-domain and cross-domain performance.
Across all training domains, accuracy drops under distribution shift—for example, from 42.05 to 29.77 in Mechanical, from 42.82 to 27.21 in Biology, and from 36.28 to 27.87 in Scientific—corresponding to decreases of roughly 8–16pp across domains.
Adding SSL on non-anonymized data (\textbf{SFT + SSL}) does not substantially improve cross-domain performance, with scores remaining similar (e.g., Mechanical 29.36, Biology 28.57).
These results suggest that conventional SSL, when applied to original headers, may preserve domain-specific lexical associations.

\paragraph{Anonymized SSL improves domain transfer.}
Introducing header anonymization during SSL yields consistent cross-domain improvements, while in-domain changes remain modest.
With \textbf{SFT + SSL + Anonymization}, cross-domain accuracy increases across all training domains:
from 29.36 to 36.80 in Mechanical, 28.57 to 30.89 in Biology, 31.91 to 36.84 in Legal, 
and 30.04 to 36.11 in Scientific.
This pattern indicates that anonymization reduces reliance on lexical cues specific to the source domain and encourages more domain-agnostic reasoning.

\paragraph{Sketches as structural supervision.}
To isolate the impact of operation sketches as a training signal, we compare \textbf{SFT + Sketch Inference} (sketches as test-time prompts) against \textbf{TaNOS w/o SSL} (sketches as training targets).
Using sketches for supervision yields clear improvements in both in-domain and cross-domain performance compared to using them only as prompts at inference time.
For instance, in the Mechanical domain, \textbf{In-domain} accuracy improves from 48.08\% to 66.79\% (+18.7pp).
\textbf{Cross-domain} performance shows an even larger gain, from 40.00\% to 66.08\% (+26.1pp).
This dual improvement suggests that explicitly \textit{learning} structural reasoning paths helps the model capture underlying reasoning patterns, leading to better task performance and improved robustness under distribution shift (see Appendix Table~\ref{tab:full-cross-domain-results}).

\paragraph{Combined effect: stable cross-domain performance.}
Combining anonymized SSL with sketch-based supervision yields the most consistent cross-domain performance among our configurations.
Across all training domains, \textbf{TaNOS} attains the highest average accuracy
(e.g., Mechanical 68.48, Biology 66.97, Legal 66.98, Scientific 67.97),
while the gap between in-domain and cross-domain performance is reduced to within about 2pp.
This pattern suggests that anonymized SSL and sketch-based supervision play complementary roles, with anonymization reducing residual lexical dependencies and sketches providing additional structural guidance, resulting in more stable performance under domain shift.

Taken together, these findings indicate that sketches provide a useful structural learning signal for program-level reasoning, while anonymized SSL encourages more domain-general representations.
In combination, these components help TaNOS generalize more reliably across diverse expert domains and outperform SFT baselines under distribution shift.
This pattern also extends to a smaller RoBERTa backbone, with even more pronounced gains (Appendix Table~\ref{tab:roberta-cross-domain-results}).


\begin{table}[htbp]
  \centering
  \small
  \begin{adjustbox}{max width=\columnwidth}
  \begin{tabular}{c|cc}
    \Xhline{1.2pt}
    & \multicolumn{2}{c}{\textbf{Test Setting}} \\
    \cline{2-3}
    \textbf{Train Setting} & Original & Anonymized\\
    \hline
    Off-the-shelf & 15.87 & 14.44 \\
    \hline
    Original     & \underline{99.54} & 83.21 \\
    Anonymized   & 99.42 & \underline{\textbf{99.25}} \\
    \Xhline{1.2pt}
  \end{tabular}
  \end{adjustbox}
    \caption{
    Program accuracy (\%) on \textbf{NumReason-500 under header anonymization} (inference-only baseline). Original-header training drops under anonymized evaluation, whereas anonymized-header training stays stable across both.
    }
  \label{tab:anonymization}
\end{table}

\subsection{Effectiveness of Anonymization}
\label{sec:anonymization-effect}

Table~\ref{tab:anonymization} examines the effect of header anonymization on lexical robustness.
To ensure that improvements stem from generalization rather than token overlap, the anonymization tokens used during training and testing are drawn from disjoint sets.
Models trained on the original headers achieve near-perfect accuracy on the matched test set (99.54),
but their performance drops to 83.21 when evaluated under anonymized headers, suggesting a sensitivity to memorized lexical cues.
In contrast, models trained with anonymized headers maintain consistently high accuracy across both evaluation settings 
(99.42 on original, 99.25 on anonymized).
Taken together, these results indicate that anonymization reduces reliance on specific headers and acts as a regularization strategy that improves robustness to lexical variation.

\subsection{Robustness to Training Data Scale}
\label{sec:data-scale}
\begin{figure}[t]
  \centering
  \includegraphics[width=0.98\columnwidth]{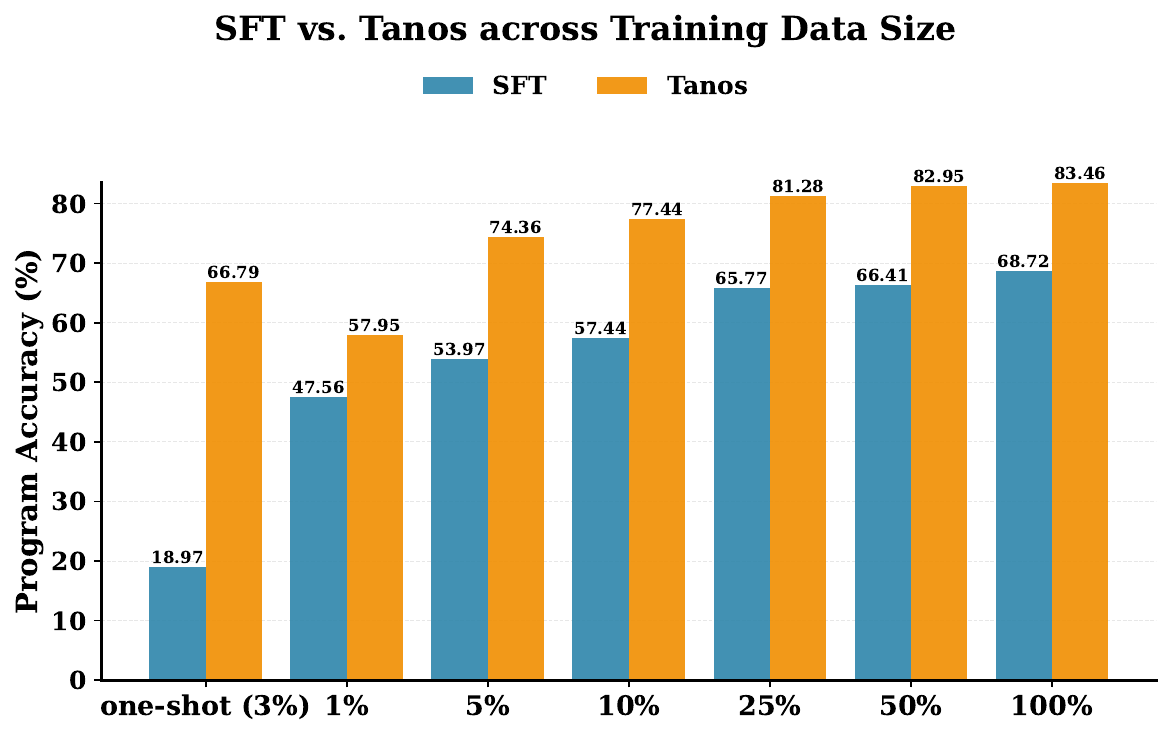}
    \caption{
      Program accuracy (\%) of TaNOS and the SFT baseline across \textbf{varying fractions of Financial TableQA} training data.
      TaNOS consistently surpasses SFT, with particularly strong gains in the \textbf{one-shot training} setting, indicating stronger pattern-level generalization.
    }
      \label{fig:data-scale}
\end{figure}

We analyze how TaNOS scales with the amount of supervision by comparing it to the SFT baseline across a wide range of data sizes (Figure~\ref{fig:data-scale}).
We observe three trends across different data regimes.

\paragraph{One-shot Training per Operation Pattern.}
In this setting, training is constrained to exactly one example for each operation pattern, yielding a \textbf{one-shot training} regime at the pattern level.
Under this supervision, SFT attains 18.97\% accuracy, indicating poor pattern generalization from such limited exposure.
In contrast, TaNOS achieves 66.79\% in the same setting, surpassing the baseline trained on 50\% of the data (66.41\%) and approaching the baseline trained on 100\% (68.72\%),
demonstrating robust pattern-level generalization from a single example per pattern.

\paragraph{Low- to mid-resource regimes.}
With 1–10\% of the data, TaNOS attains higher accuracy than the baseline (approximately 10pp at 1\%, 20pp at 5\%, and 20pp at 10\%).
This gap indicates improved sample efficiency and the ability to recover strong program accuracy from limited labeled examples.

\paragraph{High-resource regimes.}
Even with 25–100\% of the data, TaNOS maintains an advantage of around 15pp over the SFT.
Accuracy increases smoothly as data grows, suggesting that the model continues to benefit from sketch-level structural guidance.

\begin{table}[htbp]
\centering
\small
\begin{adjustbox}{max width=\columnwidth}
\begin{tabular}{l c c c}
\toprule
\textbf{Method} & \textbf{Original} & \textbf{Paraphrased} & \textbf{Exec. Para.} \\
\midrule
SFT + Sketch w train \& infer (10\%)            & 74.49 & 69.62 & 72.56 \\
TaNOS w/o Anony (10\%)          & 75.38 & 71.67 & 74.36 \\
TaNOS (10\%)                    & 77.44 & 72.95 & 75.90 \\
TaNOS (100\%)                   & \textbf{83.46} & \textbf{75.90} & \textbf{78.97}\\
\bottomrule
\end{tabular}
\end{adjustbox}
\caption{
Program accuracy (\%) on performance under \textbf{paraphrased operation sketches} (BLEU = 0.2691).
Paraphrases generated by GPT-5 introduce noticeable lexical and syntactic variation.
Execution accuracy under paraphrased sketches is also reported.
}
\label{tab:paraphrase_desc}
\end{table}

\subsection{Robustness to Sketch Variations}
\label{sec:paraphrase_desc}

To evaluate sensitivity to the surface form of operation sketches, we paraphrased all sketches using GPT-5, yielding noticeable lexical and syntactic variation (BLEU = 0.2691).
The results in Table~\ref{tab:paraphrase_desc} show that accuracy decreases for all configurations, but performance remains relatively high, indicating that the models do not rely solely on the exact phrasing of the sketches.

More extensive supervised training is associated with greater sensitivity to paraphrasing:
\textbf{TaNOS (Full)} drops from 83.46\% to 75.90\% (-7.6pp), whereas the 10\%-data variant declines more modestly (-4.5pp).
This pattern suggests that extensive fine-tuning increases dependence on specific linguistic realizations.

Systems leveraging SSL and header anonymization are more robust to paraphrased operation sketches. Even when sketches are provided during both training and inference,
removing both SSL and anonymization (SFT + Sketch) yields the lowest program accuracy (69.62\%), while removing only anonymization (TaNOS w/o Anony) improves robustness (71.67\%).
The full \textbf{TaNOS} model achieves the best performance (72.95\%),
indicating that \textbf{SSL} and \textbf{header anonymization} play complementary roles
in interpreting sketches beyond their surface lexical form.

\subsection{Generalization to Human-Curated Biology Data}
\label{sec:bio_human}

To evaluate cross-domain robustness beyond synthetic data, we assess human-curated biology dataset containing 122 expert-authored questions (Table~\ref{tab:bio_human}).
These questions introduce new linguistic constructions, reasoning styles, and table layouts that are not present in the training.

\begin{table}[t]
\centering
\small
\begin{tabular}{l cc}
\toprule
\textbf{Method} & \textbf{10\% Data} & \textbf{100\% Data} \\
\midrule
SFT                                    & 57.38 & 62.30 \\
SFT + SSL                              & 56.56 & 61.48 \\
TaNOS w/o Sketch                       & 58.20 & 59.84 \\
TaNOS w/o SSL                          & 62.18 & 68.85 \\
\textbf{TaNOS}                         & \textbf{63.85} & \textbf{70.26} \\
\bottomrule
\end{tabular}
\caption{
Program accuracy (\%) on the \textbf{expert-curated Biology dataset}.
Models are trained on a finance-domain dataset and evaluated on expert-authored biology questions.
TaNOS achieves the highest accuracy under both supervision budgets, indicating improved transfer to non-synthetic data.
}
\label{tab:bio_human}
\end{table}

\paragraph{Limited effects of anonymization and SSL under structural divergence.}
SFT yields moderate performance (57.38\% with 10\% data, 62.30\% with 100\%), and adding SSL provides no improvement (56.56\% and 61.48\%).
\textbf{TaNOS w/o Sketch}, which applies anonymized SSL without sketch guidance, also underperforms the baseline at 100\% data (59.84\% vs.\ 62.30\%).
This suggests that when table structures differ substantially from the pretraining distribution, header anonymization and SSL offer only limited benefits by themselves.

\paragraph{Sketches enable high-level contextual reasoning.}
In contrast, sketch-guided supervision remains effective under substantial structural divergence.
\textbf{TaNOS w/o SSL} surpasses SFT by 4.8pp at 10\% data and 6.6pp at 100\%, indicating that sketches guide the model beyond surface-level cues toward higher-level contextual reasoning.
The full \textbf{TaNOS} model further improves accuracy (by 6.5pp and 8.0pp over SFT), showing that anonymization and SSL become most useful when combined with sketch supervision.

\subsection{Generalization to Structurally Complex Multi-Table Reasoning}
\label{sec:multihiertt}

The preceding experiments focus on relatively simple table structures.
To evaluate whether TaNOS extends to more complex layouts, we assess it on MultiHiertt, which requires reasoning across multiple nested tables (Table~\ref{tab:multihiertt}).

\begin{table}[t]
\centering
\small
\begin{tabular}{l cc}
\toprule
\textbf{Method} & \textbf{10\% Data} & \textbf{100\% Data} \\
\midrule
SFT                             & 51.43 & 61.96 \\
SFT + SSL                       & 55.27 & 61.96 \\
TaNOS w/o Sketch                & 54.77 & 63.44 \\
TaNOS w/o SSL                   & 68.28 & 82.65 \\
\textbf{TaNOS}                  & \textbf{70.63} & \textbf{82.90} \\
\bottomrule
\end{tabular}
\caption{
Program accuracy (\%) on the \textbf{MultiHiertt benchmark}, which requires hierarchical multi-table numerical reasoning.
TaNOS achieves the highest accuracy under both supervision budgets, indicating that it extends to complex tables.
}
\label{tab:multihiertt}
\end{table}

\paragraph{Larger gains on complex structures.}
The trends observed on Financial TableQA carry over to MultiHiertt, but with larger relative gains. SFT achieves 61.96\% program accuracy with 100\% data. In contrast, \textbf{TaNOS} reaches \textbf{82.90\%}, a gain of roughly 21pp over the baseline, which exceeds the corresponding improvement on Financial TableQA (+14pp; Table~\ref{tab:appendix-program-acc}).

\section{Conclusion}

We introduced \textbf{TaNOS}, a framework for numerical reasoning over expert-domain tables that is designed to decouple surface-level lexical patterns from underlying computational structure.
The framework combines header anonymization, operation sketches, and correctness-guaranteed self-supervised pretraining to provide more structurally grounded supervision for program generation. Experiments across multiple table benchmarks indicate that TaNOS can reduce performance degradation under domain shift while maintaining competitive in-domain performance. These results suggest that lightweight structural abstractions, when paired with appropriate pretraining, may offer a practical path toward more generalizable numerical reasoning over expert-domain tables. We view this work as an initial step toward closer integration between symbolic structure and neural models, and hope it can serve as a basis for further investigation into structurally grounded numerical reasoning.

\section{Limitations}
TaNOS has several limitations.
First, the automatic data generation process assumes well-structured tables, and its effectiveness may diminish when applied to highly irregular, noisy, or layout-dependent tabular formats. We do not evaluate such settings in this work. Second, TaNOS currently assumes that, at inference time, users may provide lightweight operation sketches that convey minimal computational intent.
This assumption can hold in simple reasoning scenarios, but automating sketch generation—for example, via a small auxiliary model—remains an important direction for future work, especially in settings where user-provided hints are unavailable.

\section*{Acknowledgments}
This work was supported in part by the National Research Foundation of Korea (NRF) grant (RS-2023-00280883, RS-2023-00222663); by the National Research Foundation, Korea, under project BK21 FOUR (Dept.\ of Data Science, SNU, No.\ 5199990914569); by the Korea Institute of Science and Technology Information (KISTI) in 2026 (No.\ (KISTI)K26L3M1C1), aimed at developing KONI (KISTI Open Neural Intelligence), a large language model specialized in science and technology; and by the Institute of Information \& communications Technology Planning \& Evaluation (IITP) grant funded by the Korea government (MSIT) (RS-2025-02263754, Human-Centric Embodied AI Agents with Autonomous Decision-Making); by grant (25202MFDS003) from Ministry of Food and Drug Safety in 2025; by AI-BIO Research Grant through Seoul National University; Institute of Information \& communications Technology Planning \& Evaluation (IITP) grant funded by the Korea government (MSIT) (No.\ RS-2025-25442149, LG AI STAR Talent Development Program for Leading Large-Scale Generative AI Models in the Physical AI Domain).

\bibliography{custom}

@article{chen2021finqa,
      title={FinQA: A Dataset of Numerical Reasoning over Financial Data}, 
      author={Zhiyu Chen and Wenhu Chen and Charese Smiley and Sameena Shah and Iana Borova and Dylan Langdon and Reema Moussa and Matt Beane and Ting-Hao Huang and Bryan Routledge and William Yang Wang},
      journal={arXiv preprint arXiv:2109.00122},
      year={2021}
}

@article{zhu2021tat,
  title={TAT-QA: A question answering benchmark on a hybrid of tabular and textual content in finance},
  author={Zhu, Fengbin and Lei, Wenqiang and Huang, Youcheng and Wang, Chao and Zhang, Shuo and Lv, Jiancheng and Feng, Fuli and Chua, Tat-Seng},
  journal={arXiv preprint arXiv:2105.07624},
  year={2021}
}

@article{zhao2022multihiertt,
  title={MultiHiertt: Numerical reasoning over multi hierarchical tabular and textual data},
  author={Zhao, Yilun and Li, Yunxiang and Li, Chenying and Zhang, Rui},
  journal={arXiv preprint arXiv:2206.01347},
  year={2022}
}

@article{liu2023evaluating,
  title={Evaluating the logical reasoning ability of chatgpt and gpt-4},
  author={Liu, Hanmeng and Ning, Ruoxi and Teng, Zhiyang and Liu, Jian and Zhou, Qiji and Zhang, Yue},
  journal={arXiv preprint arXiv:2304.03439},
  year={2023}
}

@article{chen2022convfinqa,
  title={Convfinqa: Exploring the chain of numerical reasoning in conversational finance question answering},
  author={Chen, Zhiyu and Li, Shiyang and Smiley, Charese and Ma, Zhiqiang and Shah, Sameena and Wang, William Yang},
  journal={arXiv preprint arXiv:2210.03849},
  year={2022}
}

@article{devlin2018bert,
  title={Bert: Pre-training of deep bidirectional transformers for language understanding},
  author={Devlin, Jacob and Chang, Ming-Wei and Lee, Kenton and Toutanova, Kristina},
  journal={arXiv preprint arXiv:1810.04805},
  year={2018}
}

@article{amini2019mathqa,
  title={Mathqa: Towards interpretable math word problem solving with operation-based formalisms},
  author={Amini, Aida and Gabriel, Saadia and Lin, Shanchuan and Koncel-Kedziorski, Rik and Choi, Yejin and Hajishirzi, Hannaneh},
  journal={arXiv preprint arXiv:1905.13319},
  year={2019}
}

@article{dua2019drop,
  title={DROP: A reading comprehension benchmark requiring discrete reasoning over paragraphs},
  author={Dua, Dheeru and Wang, Yizhong and Dasigi, Pradeep and Stanovsky, Gabriel and Singh, Sameer and Gardner, Matt},
  journal={arXiv preprint arXiv:1903.00161},
  year={2019}
}

@article{herzig2020tapas,
  title={TaPas: Weakly supervised table parsing via pre-training},
  author={Herzig, Jonathan and Nowak, Pawe{\l} Krzysztof and M{\"u}ller, Thomas and Piccinno, Francesco and Eisenschlos, Julian Martin},
  journal={arXiv preprint arXiv:2004.02349},
  year={2020}
}

@article{yin2020tabert,
  title={TaBERT: Pretraining for joint understanding of textual and tabular data},
  author={Yin, Pengcheng and Neubig, Graham and Yih, Wen-tau and Riedel, Sebastian},
  journal={arXiv preprint arXiv:2005.08314},
  year={2020}
}

@article{chen2020hybridqa,
  title={Hybridqa: A dataset of multi-hop question answering over tabular and textual data},
  author={Chen, Wenhu and Zha, Hanwen and Chen, Zhiyu and Xiong, Wenhan and Wang, Hong and Wang, William},
  journal={arXiv preprint arXiv:2004.07347},
  year={2020}
}

@article{chen2022large,
  title={Large language models are few (1)-shot table reasoners},
  author={Chen, Wenhu},
  journal={arXiv preprint arXiv:2210.06710},
  year={2022}
}

@article{li2023chatgpt,
  title={Are chatgpt and gpt-4 general-purpose solvers for financial text analytics? an examination on several typical tasks},
  author={Li, Xianzhi and Zhu, Xiaodan and Ma, Zhiqiang and Liu, Xiaomo and Shah, Sameena},
  journal={arXiv preprint arXiv:2305.05862},
  year={2023}
}

@article{chen2022program,
  title={Program of thoughts prompting: Disentangling computation from reasoning for numerical reasoning tasks},
  author={Chen, Wenhu and Ma, Xueguang and Wang, Xinyi and Cohen, William W},
  journal={arXiv preprint arXiv:2211.12588},
  year={2022}
}

@article{liu2024augment,
  title={Augment before You Try: Knowledge-Enhanced Table Question Answering via Table Expansion},
  author={Liu, Yujian and Ji, Jiabao and Yu, Tong and Rossi, Ryan and Kim, Sungchul and Zhao, Handong and Sinha, Ritwik and Zhang, Yang and Chang, Shiyu},
  journal={arXiv preprint arXiv:2401.15555},
  year={2024}
}

@article{geva2020injecting,
  title={Injecting numerical reasoning skills into language models},
  author={Geva, Mor and Gupta, Ankit and Berant, Jonathan},
  journal={arXiv preprint arXiv:2004.04487},
  year={2020}
}

@inproceedings{ran2019numnet,
  title     = {NumNet: Machine Reading Comprehension with Numerical Reasoning},
  author    = {Ran, Qiu and Lin, Yankai and Li, Peng and Zhou, Jie and Liu, Zhiyuan},
  booktitle = {Proceedings of the 2019 Conference on Empirical Methods in Natural Language Processing and the 9th International Joint Conference on Natural Language Processing (EMNLP-IJCNLP)},
  year      = {2019}
}

@inproceedings{zhang2022elastic,
  title     = {{ELASTIC}: Numerical Reasoning with Adaptive Symbolic Compiler},
  author    = {Zhang, Jiaxin and Moshfeghi, Yashar},
  booktitle = {Advances in Neural Information Processing Systems},
  year      = {2022}
}

@article{dubey2024llama,
  title={The llama 3 herd of models},
  author={Dubey, Abhimanyu and Jauhri, Abhinav and Pandey, Abhinav and Kadian, Abhishek and Al-Dahle, Ahmad and Letman, Aiesha and Mathur, Akhil and Schelten, Alan and Yang, Amy and Fan, Angela and others},
  journal={arXiv e-prints},
  pages={arXiv--2407},
  year={2024}
}

@misc{qwen2025qwen25technicalreport,
      title={Qwen2.5 Technical Report}, 
      author={Qwen and : and An Yang and Baosong Yang and Beichen Zhang and Binyuan Hui and Bo Zheng and Bowen Yu and Chengyuan Li and Dayiheng Liu and Fei Huang and Haoran Wei and Huan Lin and Jian Yang and Jianhong Tu and Jianwei Zhang and Jianxin Yang and Jiaxi Yang and Jingren Zhou and Junyang Lin and Kai Dang and Keming Lu and Keqin Bao and Kexin Yang and Le Yu and Mei Li and Mingfeng Xue and Pei Zhang and Qin Zhu and Rui Men and Runji Lin and Tianhao Li and Tianyi Tang and Tingyu Xia and Xingzhang Ren and Xuancheng Ren and Yang Fan and Yang Su and Yichang Zhang and Yu Wan and Yuqiong Liu and Zeyu Cui and Zhenru Zhang and Zihan Qiu},
      year={2025},
      archivePrefix={arXiv},
      primaryClass={cs.CL},
}

@article{comanici2025gemini,
  title={Gemini 2.5: Pushing the frontier with advanced reasoning, multimodality, long context, and next generation agentic capabilities},
  author={Comanici, Gheorghe and Bieber, Eric and Schaekermann, Mike and Pasupat, Ice and Sachdeva, Noveen and Dhillon, Inderjit and Blistein, Marcel and Ram, Ori and Zhang, Dan and Rosen, Evan and others},
  journal={arXiv preprint arXiv:2507.06261},
  year={2025}
}

@article{openai2025gpt5,
  title   = {{OpenAI GPT-5} System Card},
  author  = {Singh, Aaditya and Fry, Adam and Perelman, Adam and Tart, Adam and others},
  journal = {arXiv preprint arXiv:2601.03267},
  year    = {2025}
}
\bibliographystyle{acl_natbib}
\appendix

\section{Implementation Details}
\label{app:impl}

This appendix summarizes the set of implementation details required to
reproduce our experiments. All models were fine-tuned using the HuggingFace
\texttt{transformers} library, and we report only the components that directly
affect training dynamics and evaluation.

\subsection{Chat Prompt Format}

We adopt a chat-style interface for all open-weight models.
The system message is fixed as:
\begin{itemize}[leftmargin=1.2em,itemsep=0.2em,topsep=0.2em]
    \item \textbf{System}: ``You are an assistant that returns only the math program solving the user request. Respond with the program text only.''
\end{itemize}

For each training example, the user turn is constructed by concatenating:
(i) the question \texttt{qa["question"]},
(ii) the linearized table and textual context \texttt{qa["model\_input"]},
and (optionally) (iii) the gold sketch \texttt{qa["explanation\_math"]},
separated by a literal \texttt{[SEP]} token:
\[
\text{user} = \text{question} \; [\texttt{SEP}] \; \text{context} \; [\texttt{SEP}] \; \text{sketch}.
\]

We rely on the built-in \texttt{chat\_template} of each tokenizer to render the
(System, User, Assistant) turns whenever available; otherwise, we fall back to a
simple textual prefix (``System:'', ``User:'', ``Assistant:'').

During supervised training, the gold program is appended as the assistant turn.
All tokens preceding the assistant prefix are masked with label \(-100\), so that
the model is trained \emph{only} on the program tokens.

\subsection{Training Hyperparameters}

All supervised fine-tuning experiments (TaNOS, TaNOS w/o SSL, and SFT baselines) use the following configuration:
\begin{itemize}[leftmargin=1.2em,itemsep=0.2em,topsep=0.2em]
    \item Backbone:
    \begin{itemize}[leftmargin=1.5em,itemsep=0.1em,topsep=0.1em]
        \item \texttt{LLaMA-3.1-8B-Instruct}
        \item \texttt{LLaMA-3.2-3B-Instruct}
        \item \texttt{Qwen-2.5-7B-Instruct}
        \item \texttt{Qwen-2.5-3B-Instruct}
    \end{itemize}
    \item Optimizer: AdamW
    \item Learning rate: \(1\times10^{-5}\)
    \item Epochs: 10
    \item Effective batch size: 16 (batch size 4, gradient accumulation 4)
    \item Max sequence length: 1024
    \item LR schedule: constant LR with 3\% warmup
    \item Precision: \texttt{bf16}
    \item Gradient checkpointing: enabled
    \item Random seed: 42
    \item Model selection: unless otherwise specified, we select the checkpoint
    achieving the highest program accuracy on the dev set and report
    its test performance.
\end{itemize}

\paragraph{SSL pretraining.}
For self-supervised learning (SSL), we use the same configuration except for:
\begin{itemize}[leftmargin=1.2em,itemsep=0.2em,topsep=0.2em]
    \item \textbf{Epochs}: 1
    \item \textbf{Scheduler}: cosine LR schedule
\end{itemize}

\subsection{Program Generation and Decoding}

At evaluation time (both for development-time model selection and final test
evaluation), we generate programs with greedy decoding:
\begin{itemize}[leftmargin=1.2em,itemsep=0.2em,topsep=0.2em]
    \item \textbf{Decoding}: \texttt{do\_sample = False}
    \item \textbf{Max new tokens}: 128
    \item \textbf{Termination}: model EOS token
\end{itemize}
We always discard the prompt portion of the sequence and keep only tokens
generated \emph{after} the final user turn.
For batched generation, tokenizers use right padding during training and left
padding during evaluation for efficiency, with the pad token tied to the EOS
token.

\subsection{Metrics: Program and Execution Accuracy}

\paragraph{Program accuracy.}
To compute program accuracy, we post-process the generated text as follows:
(i) we take the substring after the last occurrence of the token
``assistant'' in the decoded text;
(ii) we strip leading punctuation and whitespace; and
(iii) we select the first non-empty line as the predicted program.
We then compute exact string match between this line and the gold program.
All reported program accuracies in the paper are based on this strict exact
match criterion.

\paragraph{Execution accuracy.}
For execution accuracy, we evaluate generated programs with a custom interpreter
that implements the arithmetic and table operations used in FinQA (e.g.,
\texttt{add}, \texttt{subtract}, \texttt{multiply}, \texttt{divide},
\texttt{table\_sum}, \texttt{table\_average}, \texttt{table\_max},
\texttt{table\_min}).
The interpreter executes the sequence of operations and returns a scalar numeric
answer.
To decide correctness, we apply a FinQA-style numeric matching rule:
we first normalize both the predicted and gold answers by lowercasing, removing
commas, currency symbols, and unit markers (e.g., ``million'', ``k''),
and stripping extraneous punctuation and whitespace.
We then interpret simple numeric forms such as percentages (with or without
the ``\%'' sign), ratios and fractions (e.g., ``3/4''), and map
\texttt{yes}/\texttt{no}-style outputs to booleans.
If both answers contain numeric values, we require them to have the same sign
and to agree after rounding to the number of decimal places specified in either
the gold answer or the prediction.
We do not apply any additional numeric tolerance beyond this rounding-based
equivalence.

\subsection{Prompt for Non-finetuned Model Evaluation}

For all non-finetuned models---including API-based LLMs
(e.g., \textbf{GPT-5}, \textbf{GPT-5-Nano}, \textbf{Claude-4.5-Sonnet},
\textbf{Claude-4.5-Haiku}, \textbf{Gemini-2.5-Pro}, \textbf{Gemini-2.5-Flash-Lite})
and open-weight instruction-tuned models evaluated in the inference-only setting
(e.g., \textbf{LLaMA-3.3-70B-Instruct}, \textbf{Qwen-2.5-72B-Instruct},
\textbf{LLaMA-3.1-8B-Instruct}, \textbf{LLaMA-3.2-3B-Instruct},
\textbf{Qwen-2.5-7B-Instruct}, \textbf{Qwen-2.5-3B-Instruct})---we use a
shared instruction-style prompt:
\begin{quote}
\small
\texttt{Based on the provided context and table, answer the question.}

\texttt{Context:}

\texttt{\{context\}}

\texttt{Table:}

\texttt{\{table\}}

\texttt{Question: \{question\}}

\texttt{Important: Provide only the exact numeric answer. Do not include any explanation or additional information.}
\end{quote}
Here, \texttt{\{context\}} is formed by concatenating the pre- and post-text
segments from the FinQA instance, and \texttt{\{table\}} is a row-wise
linearization of the table.
When evaluating the sketch condition, we append an additional block:
\begin{quote}
\small
\texttt{Sketch:}

\texttt{\{explanation\_math\}}
\end{quote}
Model outputs are treated as numeric answers and evaluated with the same
normalization and rounding-based numeric matching rule as described in the
\emph{Execution accuracy} subsection.

\subsection{Hardware and Quantization}

All experiments were executed on a single \textbf{NVIDIA H100~SXM (80GB) GPU}.
We enable deterministic CUDA behavior (\texttt{torch.use\_deterministic\_algorithms})
and disable cuDNN benchmarking to improve reproducibility.

Large open-weight models (LLaMA-3.3-70B-Instruct and Qwen-2.5-72B-Instruct)
are loaded with 4-bit NF4 quantization using the bitsandbytes library, with
\texttt{bfloat16} computation and \texttt{device\_map="auto"} for efficient
inference.
All open-weight models used in our experiments are instruction-tuned variants.
\clearpage
\section{NumReason-500 Jargon and Examples}
\label{app:numreason-examples}

\begin{center}

\begin{minipage}{\textwidth}
\centering
\setlength{\tabcolsep}{6pt}
\begin{adjustbox}{max width=\textwidth}
\begin{tabular}{c|l}
\toprule
\textbf{Key} & \textbf{Jargon} \\
\midrule\midrule
\texttt{operation\_0} &
\makecell[l]{\texttt{goodwill} + \texttt{intangibleAssets} \\
            - \texttt{otherFinancingActivities} + \texttt{minorityInterest}} \\
\midrule
\texttt{operation\_1} &
\makecell[l]{\texttt{commonStock} + \texttt{commonStock}} \\
\midrule
\texttt{operation\_2} &
\makecell[l]{\texttt{const\_100} $\times$ [\texttt{numberOfShares} - \texttt{costAndExpenses}]} \\
\bottomrule
\end{tabular}
\end{adjustbox}
\captionof{table}{Examples of operation-level jargon definitions in NumReason-500.}
\label{tab:jargon-list}
\end{minipage}

\vspace{2mm}

\begin{minipage}{\textwidth}
\centering
\setlength{\tabcolsep}{6pt}
\begin{adjustbox}{max width=\textwidth}
\begin{tabular}{l}
\toprule
\textbf{Examples} \\
\midrule\midrule

\textbf{Question}: What is the \textit{combined amount of} {\color{magenta}other financing activities}
in Q3 and the full-year {\color{magenta}operating cash flow}? \\
\textbf{Sketch}: $ {\color{magenta}\text{Other Financing Activities}} + {\color{magenta}\text{Operating Cash Flow}} $ \\
\textbf{Program}: \texttt{add(-3000000.0, 1697000000.0)} \\
\textbf{Answer}: $1694000000.0$ \\
\midrule

\textbf{Question}: What is the \textit{percentage change} in {\color{magenta}operating cash flow}
from Q1 to Q2? \\
\textbf{Sketch}: $ \big({\color{magenta}\text{Operating Cash Flow}} - {\color{magenta}\text{Operating Cash Flow}}\big) / {\color{magenta}\text{Operating Cash Flow}}$ \\
\textbf{Program}: \texttt{subtract(85900000.0, 82100000.0), divide(\#0, 82100000), multiply(\#1, 100)} \\
\textbf{Answer}: $4.63$ \\
\midrule

\textbf{Question}: What is DVN's Q4 2021 {\color{teal}\texttt{operation\_2}}? \\
\textbf{Sketch}: \texttt{const\_100} $\times$ [\,{\color{magenta}Number Of Shares} - {\color{magenta}Cost And Expenses}\,] \\
\textbf{Program}: \texttt{subtract(671000, 257600), multiply(const\_100, \#0)} \\
\textbf{Answer}: $41340000.0$ \\
\midrule

\textbf{Question}: What is SWKS's 2016 {\color{teal}\texttt{operation\_1}} from Q3 to Q4? \\
\textbf{Sketch}: {\color{magenta}Common Stock} + {\color{magenta}Common Stock} \\
\textbf{Program}: \texttt{add(46900000, 46900000)} \\
\textbf{Answer}: $93800000.0$ \\
\midrule

\textbf{Question}: What is DLTR's Q4 2007 {\color{teal}Return On Equity}? \\
\textbf{Sketch}: {\color{magenta}Net Income} / {\color{magenta}Total Stockholders Equity} \\
\textbf{Program}: \texttt{divide(97600000, 1167700000)} \\
\textbf{Answer}: $0.0835$ \\
\bottomrule
\end{tabular}
\end{adjustbox}
\captionof{table}{
Illustrative examples from the NumReason-500 dataset before anonymization.
Operation identifiers (jargon) are highlighted in {\color{teal}teal}, and row headers are shown in {\color{magenta}magenta}.
The last example corresponds to a standard financial ratio (\emph{return on equity}, ROE).
After anonymization, the question text alone no longer reveals this underlying financial concept,
so the model must rely on the sketch to recover the correct program.
}
\label{tab:numreason-examples}
\end{minipage}

\end{center}
\newpage
\clearpage
\section{Full Experiment Results}
\label{app:full-results}


\subsection{Cross-Dataset Generalization}

\begin{center}
\begin{minipage}{\textwidth}
\centering
\setlength{\tabcolsep}{6pt}
\begin{adjustbox}{max width=\textwidth}
\begin{tabular}{l | ccc}
\toprule
 & \multicolumn{3}{c}{\textbf{Test Data}} \\
\cmidrule(lr){2-4}
\textbf{Train Data} & FinQA & MultiHiertt & NumReason-500 \\
\midrule[\heavyrulewidth]
FinQA (SFT)          & \underline{68.72} & 43.74 & 39.66 \\
MultiHiertt (SFT)    & 50.51 & \underline{61.96} & 13.60 \\
NumReason-500 (SFT)  & 27.44 & 21.56 & \underline{99.54} \\
\midrule
\textbf{FinQA (TaNOS)} & \underline{\textbf{83.46}} & \textbf{70.01} & \textbf{83.89} \\
\bottomrule
\end{tabular}
\end{adjustbox}

\captionof{table}{
Program accuracy (\%) on \textbf{Cross-Dataset Generalization}.
Rows denote training data, columns test data. In-domain results are underlined.
SFT degrades under subtle distribution shift.
\textbf{TaNOS}, pretrained on anonymized NumReason-500 and fine-tuned on FinQA, surpasses in-domain SFT on unseen MultiHiertt when provided with sketch guidance—indicating robust generalization.
}
\label{tab:cross-domain-financial}
\end{minipage}
\end{center}

\subsection{Execution Accuracy of Models}

\begin{center}
\begin{minipage}{\textwidth}
\centering
\setlength{\tabcolsep}{6pt}
\begin{adjustbox}{max width=\textwidth}
\begin{tabular}{l | cc}
\toprule
\textbf{Model} 
& \multicolumn{2}{c}{\textbf{Execution Accuracy (\%)}} \\
\cmidrule(lr){2-3}
& \ding{55} & \ding{51} \\
\midrule[\heavyrulewidth]

GPT-5                  & 71.28 & 77.82 \\
Claude-4.5-Sonnet      & 73.97 & 76.79 \\
Gemini-2.5-Pro         & 70.00 & 76.41 \\
\midrule
GPT-5-Nano             & 69.49 & 71.41 \\
Claude-4.5-Haiku       & 68.33 & 71.54 \\
Gemini-2.5-Flash-Lite  & 42.05 & 41.03 \\
\midrule
Qwen-2.5-72B-Instruct (4-bit)  & 56.54 & 60.26 \\
LLaMA-3.3-70B-Instruct (4-bit) & 50.77 & 51.92 \\
\midrule
LLaMA-3.1-8B-Instruct & 21.15 & 23.59 \\
Qwen-2.5-7B-Instruct  & 35.90 & 38.08 \\
\midrule
Qwen-2.5-3B-Instruct  & 13.21 & 12.44 \\
LLaMA-3.2-3B-Instruct &  7.56 & 10.64 \\
\bottomrule
\end{tabular}
\end{adjustbox}

\captionof{table}{
Execution accuracy (\%) on \textbf{Financial TableQA for API-based and open-weight LLMs.}
Columns report execution accuracy with (\ding{51}) and without (\ding{55}) sketch guidance at inference time.
}
\label{tab:appendix-llm-exec-acc}
\end{minipage}
\end{center}



\subsection{Program Accuracy of Small- and Mid-Scale Models}

\begin{center}
\begin{minipage}{\textwidth}
\centering
\setlength{\tabcolsep}{4pt}
\begin{adjustbox}{max width=\textwidth}
\begin{tabular}{l | cc | cc | cc | cc}
\toprule
\textbf{Small- and mid-scale models}
    & \multicolumn{2}{c|}{\textit{LLaMA-3.2-3B}}
    & \multicolumn{2}{c|}{\textit{Qwen-2.5-3B}}
    & \multicolumn{2}{c|}{\textit{LLaMA-3.1-8B}}
    & \multicolumn{2}{c}{\textit{Qwen-2.5-7B}} \\
\cmidrule(lr){1-9}
\textbf{Sketch}
    & \ding{55} & \ding{51}
    & \ding{55} & \ding{51}
    & \ding{55} & \ding{51}
    & \ding{55} & \ding{51} \\
\midrule

SFT (10\%) 
& 47.69 & 52.05
& 50.90 & 50.26
& 57.44 & 62.82
& 58.59 & \textbf{63.97} \\

SFT (100\%)
& 62.05 & 62.82
& 57.05 & 61.67
& 68.72 & \textbf{69.87}
& 66.28 & 66.54 \\
\midrule

\textbf{TaNOS (10\%)}
& 28.85 & 67.31
& 23.21 & 71.28
& 38.08 & \textbf{77.44}
& 27.95 & 76.28 \\

\textbf{TaNOS (100\%)}
& 28.08 & 79.87
& 36.28 & 79.36
& 37.05 & \textbf{83.46}
& 44.87 & 82.82 \\
\bottomrule
\end{tabular}
\end{adjustbox}

\captionof{table}{
Program Accuracy (\%) on \textbf{Financial TableQA for small- and mid-scale models.}
}
\label{tab:appendix-program-acc}
\end{minipage}
\end{center}
\newpage
\clearpage

\subsection{Full Cross-Domain Experiment Results}

\begin{center}
\begin{minipage}{\textwidth}
\centering
\footnotesize
\setlength{\tabcolsep}{4pt}
\renewcommand{\arraystretch}{1.1}

\begin{adjustbox}{max width=\textwidth}
\begin{tabular}{l l c r r r r r r r r}
\toprule
\textbf{Train Domain} &
\textbf{Method} &
\textbf{Data} &
\textbf{Avg.} &
\textbf{In-domain} &
\textbf{Cross-domain} &
\textbf{Mechanical} &
\textbf{Biology} &
\textbf{Legal} &
\textbf{Scientific} &
\textbf{Anonymized} \\
\midrule

\multirow{6}{*}{Mechanical}
& SFT
& 10\% & 31.82 & 42.05 & 29.77 & 42.05 & 34.10 & 26.79 & 31.54 & 27.95 \\
& SFT + SSL
& 10\% & 31.54 & 42.44 & 29.36 & 42.44 & 31.92 & 27.05 & 28.85 & 29.36 \\
& SFT + SSL + Anonymization
& 10\% & 37.95 & 43.72 & 36.80 & 43.72 & 37.69 & 34.87 & 37.95 & 36.92 \\
& SFT + Sketch Inference
& 10\% & 41.62 & 48.08 & 40.00 & 48.08 & 40.64 & 37.44 & 41.79 & 40.13 \\
& TaNOS w/o SSL
& 10\% & 66.20 & 66.79 & 66.08 & 66.79 & 66.15 & 64.23 & 66.54 & 66.15 \\
& \textbf{TaNOS}
& 10\% & \textbf{68.48} & \textbf{68.85} & \textbf{68.41} & \textbf{68.85} & \textbf{69.36} & \textbf{67.69} & \textbf{69.10} & \textbf{68.08} \\
\midrule

\multirow{6}{*}{Biology}
& SFT
& 10\% & 29.81 & 42.82 & 27.21 & 26.79 & 42.82 & 22.31 & 29.10 & 29.10 \\
& SFT + SSL
& 10\% & 30.60 & 40.77 & 28.57 & 29.23 & 40.77 & 25.00 & 28.85 & 28.97 \\
& SFT + SSL + Anonymization
& 10\% & 32.86 & 42.69 & 30.89 & 28.21 & 42.69 & 26.28 & 30.13 & 36.15 \\
& SFT + Sketch Inference
& 10\% & 43.61 & 51.79 & 41.57 & 41.92 & 51.79 & 36.92 & 43.85 & 43.59 \\
& TaNOS w/o SSL
& 10\% & 65.15 & 66.67 & 64.85 & 65.51 & 66.67 & 64.10 & 64.36 & 64.87 \\
& \textbf{TaNOS}
& 10\% & \textbf{66.97} & \textbf{66.79} & \textbf{67.01} & \textbf{67.56} & \textbf{66.79} & \textbf{66.03} & \textbf{67.44} & \textbf{67.31} \\
\midrule

\multirow{6}{*}{Legal}
& SFT
& 10\% & 32.93 & 40.64 & 30.72 & 31.67 & 34.36 & 40.64 & 30.64 & 32.18 \\
& SFT + SSL
& 10\% & 33.29 & 39.74 & 31.91 & 31.41 & 33.21 & 39.74 & 32.69 & 33.72 \\
& SFT + SSL + Anonymization
& 10\% & 37.76 & 42.95 & 36.84 & 36.79 & 38.59 & 42.95 & 36.41 & 36.79 \\
& SFT + Sketch Inference
& 10\% & 43.18 & 48.46 & 41.86 & 39.23 & 43.46 & 48.46 & 41.67 & 43.08 \\
& TaNOS w/o SSL
& 10\% & 66.69 & \textbf{67.05} & 66.40 & 66.41 & \textbf{67.56} & \textbf{67.05} & 65.38 & \textbf{67.31} \\
& \textbf{TaNOS}
& 10\% & \textbf{66.98} & 66.41 & \textbf{67.08} & \textbf{67.69} & 67.05 & 66.41 & \textbf{67.18} & 66.79 \\
\midrule

\multirow{6}{*}{Scientific}
& SFT
& 10\% & 30.49 & 36.28 & 27.87 & 31.28 & 31.54 & 23.72 & 36.28 & 30.64 \\
& SFT + SSL
& 10\% & 33.46 & 41.67 & 30.04 & 31.28 & 34.74 & 29.36 & 41.67 & 30.38 \\
& SFT + SSL + Anonymization
& 10\% & 37.82 & 43.72 & 36.11 & 35.77 & 38.59 & 34.62 & 43.72 & 39.10 \\
& SFT + Sketch Inference
& 10\% & 41.57 & 44.36 & 40.87 & 42.31 & 41.03 & 37.82 & 44.36 & 42.31 \\
& TaNOS w/o SSL
& 10\% & 65.24 & 67.18 & 64.48 & 65.77 & 65.51 & 62.18 & 67.18 & 65.00 \\
& \textbf{TaNOS}
& 10\% & \textbf{67.97} & \textbf{69.23} & \textbf{67.34} & \textbf{67.95} & \textbf{68.21} & \textbf{66.67} & \textbf{69.23} & \textbf{67.95} \\
\bottomrule
\end{tabular}
\end{adjustbox}

\captionof{table}{
Program accuracy (\%) on \textbf{full cross-domain performance} when training on 10\% data and evaluating on various domain-shifted variants.
The columns show the average performance (\textbf{Avg.}), performance on the source domain (\textbf{In-domain}),
the average on all other domains (\textbf{Cross-domain}), and the specific accuracy for each evaluation domain.
}
\label{tab:full-cross-domain-results}
\end{minipage}
\end{center}


\subsection{RoBERTa-based Cross-Domain Results}

\begin{center}
\begin{minipage}{\textwidth}
\centering
\footnotesize
\setlength{\tabcolsep}{4pt}
\renewcommand{\arraystretch}{1.1}
\begin{adjustbox}{max width=\textwidth}
\begin{tabular}{l l c r r r r r r}
\toprule
\textbf{Train Domain} &
\textbf{Method} &
\textbf{Data} &
\textbf{Avg.} &
\textbf{In-domain} &
\textbf{Cross-domain} &
\textbf{Mechanical} &
\textbf{Biology} &
\textbf{Anonymized} \\
\midrule

\multirow{4}{*}{Mechanical}
& SFT
& 10\% & 26.82 & 31.01 & 24.72 & 31.01 & 28.77 & 20.67 \\
& SFT + SSL + Anonymization
& 10\% & 29.28 & 35.05 & 26.39 & 35.05 & 29.47 & 23.32 \\
& TaNOS w/o SSL
& 10\% & 49.25 & 54.61 & 46.58 & 54.61 & 52.93 & 40.22 \\
& \textbf{TaNOS}
& 10\% & \textbf{65.78} & \textbf{67.32} & \textbf{65.02} & \textbf{67.32} & \textbf{63.83} & \textbf{66.20} \\
\midrule

\multirow{4}{*}{Biology}
& SFT
& 10\% & 18.81 & 23.74 & 16.34 & 20.11 & 23.74 & 12.57 \\
& SFT + SSL + Anonymization
& 10\% & 30.96 & 34.78 & 29.05 & 31.70 & 34.78 & 26.40 \\
& TaNOS w/o SSL
& 10\% & 37.76 & 43.16 & 35.05 & 42.04 & 43.16 & 28.07 \\
& \textbf{TaNOS}
& 10\% & \textbf{63.45} & \textbf{62.71} & \textbf{63.83} & \textbf{65.50} & \textbf{62.71} & \textbf{62.15} \\
\bottomrule
\end{tabular}
\end{adjustbox}

\captionof{table}{
Program accuracy (\%) on \textbf{RoBERTa-based cross-domain performance} when training on 10\% data and evaluating on various domain-shifted variants.
\textbf{Avg.} is the mean over the three evaluation domains, \textbf{In-domain} is accuracy on the train domain,
and \textbf{Cross-domain} is the average over the remaining two domains.
While SSL and anonymization provide only incremental gains on the larger LLaMA-3.1-8B-Instruct backbone,
they yield much more pronounced improvements for the smaller RoBERTa model.
Despite differences in architecture and model scale, the final TaNOS performance differs by at most 4\,pp between the two settings,
in sharp contrast to the 17--27\,pp gap observed for TaNOS w/o SSL.
}
\label{tab:roberta-cross-domain-results}
\end{minipage}
\end{center}

\end{document}